\documentclass[journal]{IEEEtran}
\usepackage{generic}
\usepackage{cite}
\usepackage{amsmath,amssymb,amsfonts}
\usepackage{algorithmic}
\usepackage{graphicx}
\usepackage{bbm}
\usepackage{bm}
\usepackage{booktabs}
\usepackage{makecell}
\usepackage{multirow}
\usepackage{marvosym}
\usepackage{textcomp}
\graphicspath{{figures/}}

\usepackage[colorlinks]{hyperref}
\usepackage[capitalise]{cleveref}

\newsavebox\CBox
\def\textBF#1{\sbox\CBox{#1}\resizebox{\wd\CBox}{\ht\CBox}{\textbf{#1}}}

\def\BibTeX{{\rm B\kern-.05em{\sc i\kern-.025em b}\kern-.08em
    T\kern-.1667em\lower.7ex\hbox{E}\kern-.125emX}}
\begin{document}
\title{Med-Query: Steerable Parsing of 9-DoF Medical Anatomies with Query Embedding}
\author{Heng Guo, Jianfeng Zhang, Ke Yan, Le Lu, \IEEEmembership{Fellow, IEEE}, and Minfeng Xu*
\thanks{H. Guo, J. Zhang, K. Yan, L. Lu and M. Xu are all with DAMO Academy, Alibaba Group (\Letter gh205191@alibaba-inc.com). }
\thanks{Corresponding author: Minfeng Xu (\Letter  eric.xmf@alibaba-inc.com).}
}

\maketitle

\begin{abstract}
Automatic parsing of human anatomies at the instance-level from 3D computed tomography (CT) is a prerequisite step for many clinical applications. The presence of pathologies, broken structures or limited field-of-view (FOV) can all make anatomy parsing algorithms vulnerable. In this work, we explore how to leverage and implement the successful detection-then-segmentation paradigm for 3D medical data, and propose a steerable, robust, and efficient computing framework for detection, identification, and segmentation of anatomies in CT scans. 
Considering the complicated shapes, sizes, and orientations of anatomies, without loss of generality, we present a nine degrees of freedom (9-DoF) pose estimation solution in full 3D space using a novel single-stage, non-hierarchical representation. Our whole framework is executed in a steerable manner where any anatomy of interest can be directly retrieved to further boost inference efficiency. We have validated our method on three medical imaging parsing tasks: ribs, spine, and abdominal organs. For rib parsing, CT scans have been annotated at the rib instance-level for quantitative evaluation, similarly for spine vertebrae and abdominal organs. Extensive experiments on 9-DoF box detection and rib instance segmentation demonstrate the high efficiency and effectiveness of our framework (with the identification rate of 97.0\% and the segmentation Dice score of 90.9\%), compared favorably against several strong baselines (e.g., CenterNet, FCOS, and nnU-Net). For spine parsing and abdominal multi-organ segmentation, our method achieves competitive results on par with state-of-the-art methods on the public CTSpine1K dataset and FLARE22 competition, respectively. Our annotations, code, and models are available at: \href{https://github.com/alibaba-damo-academy/Med\_Query}{Med-Query}.
\end{abstract}


\section{Introduction}

\begin{figure}[t]
\centering
\includegraphics[width=\linewidth]{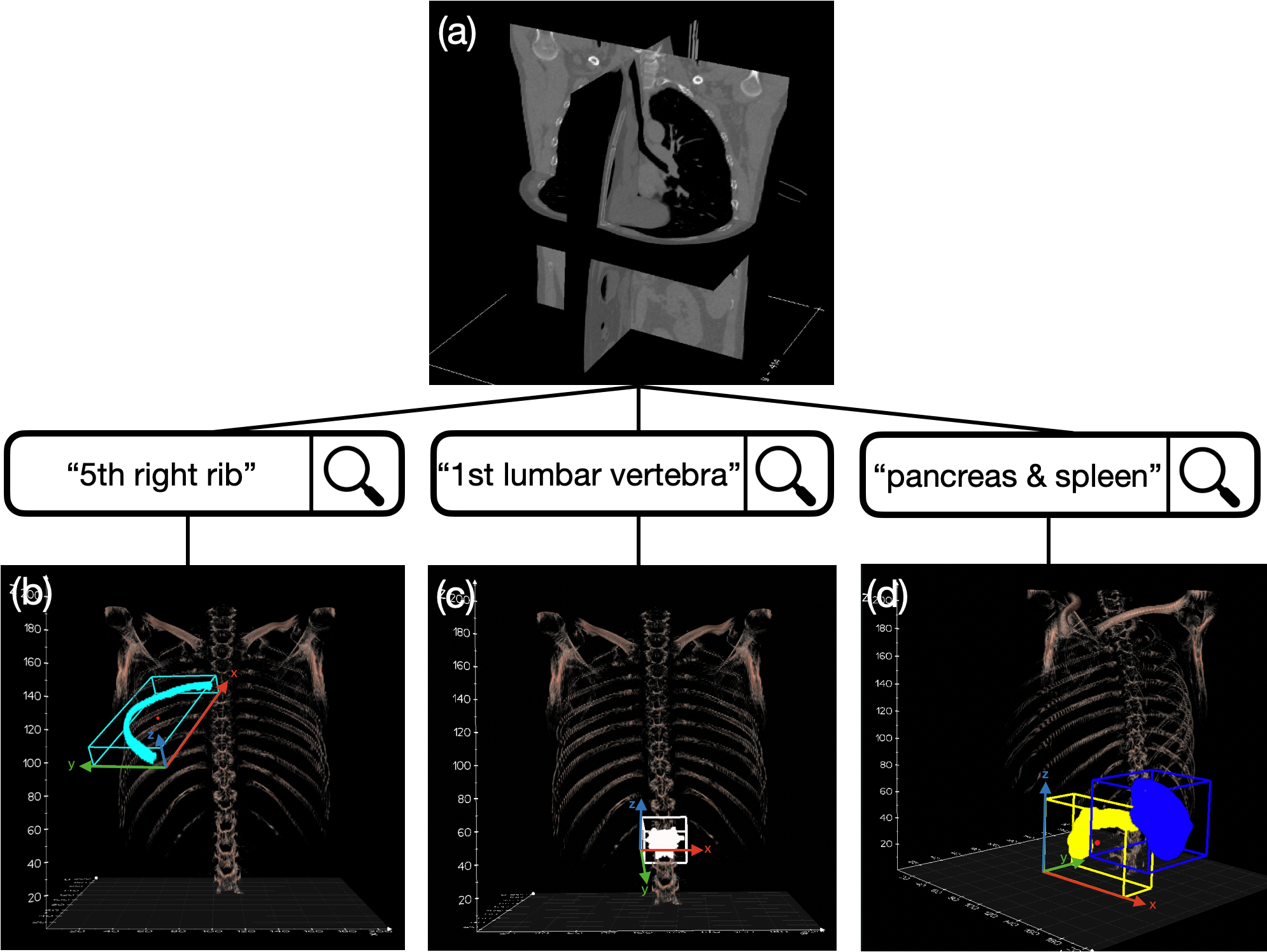}
\caption{Illustration of the steerable anatomy parsing concept. (a) Input 3D CT scan. (b) Target rib identified by query. (c) Target vertebra identified by query. (d) Identifying pancreas and spleen simultaneously. The entity in the query text will be mapped to its corresponding learned query embedding. Best viewed in color.}
\label{concept_illus}
\end{figure}

\label{sec:introduction}
Deep learning-based methods have been remarkably successful in various medical imaging tasks~\cite{chen2022recent}. 
However, current detection and segmentation algorithms often suffer from inefficiencies when applied to tasks with varying scales and complexities, particularly in scenarios where only a subset of possible structures needs to be examined. These approaches tend to process the entire image uniformly, which can result in unnecessary computations and longer inference time.
For instance, clinicians may use automatically segmented ribs around the target tumor as organs-at-risk in stereotactic body radiation therapy (SBRT)~\cite{stam2016validation}. In opportunistic screening of osteoporosis using routine CT, the first lumbar vertebra is the common anatomy of interest~\cite{jang2019opportunistic}. From the computing side, these two applications require precisely parsing ribs or vertebrae into individual instances and assigning each one its anatomical label, e.g., ``5th right rib'' or ``1st lumbar vertebra''. Extensive previous works have been proposed for nearly two decades \cite{shen2004tracing, klinder2007automated, staal2007automatic, ramakrishnan2011automatic2, wu2012learning, lenga2018deep, yang2021ribseg, yao2006automated, klinder2009automated, sekuboyina2021verse}, among them complicated heuristic parsing rules were constructed in order to obtain robust and holistically valid anatomy labeling results. 
This motivates us to develop a steerable system that enables adaptive focus on queried structures. By steering the algorithm's attention only to relevant regions, it is possible to achieve faster inference and improved computational efficiency. This advantage is particularly crucial in applications such as 3D medical imaging analysis, where rapid and accurate detection/segmentation of specific structures can significantly save time for doctors and patients. The expected system should have \textit{steerability}, which in the context of detection/segmentation algorithms, refers to allowing for adaptive processing based on the user's query. For example, if the 5th right rib is expected to be analyzed, a steerable system should output the target rib's localization and pixel-level segmentation mask if it exists in the input scan; otherwise, it should output empty results, as intuitively illustrated in~\cref{concept_illus}. 

To achieve this goal, we need an effective deep feature embedding to encode high-level anatomical information including semantic label, location, size, shape, texture, and relation with contextual circumstances. The success of Transformer in natural language processing and its application in object detection in computer vision shed light on such embedding learning~\cite{vaswani2017attention, carion2020end}. DETR, the first Transformer-based object detector, models object detection as a set prediction problem and uses bipartite matching to bind query predictions with the ground-truth boxes during training~\cite{carion2020end}. In this work, we explore to utilize some intrinsic properties in medical imaging, such as label uniqueness (e.g., there is only one ``5th right rib''), to achieve a steerable detection scheme. An effective weighted adjacency matrix is proposed to guide the bipartite matching process. Once converged, each query embedding will be assigned an anatomical semantic label, enabling the model to operate steerably. Thus, our model is named Med-Query, i.e., steerable medical anatomy parsing with query embedding.

Detection in full 3D space is non-trivial, especially for repeated anatomical sequential structures, such as ribs or vertebrae. For rib parsing, the first barrier that we have to overcome is how to represent each elongated and oblique rib using a properly parameterized bounding box. The ordinary axis-aligned 3D bounding boxes enclosing ribs will have large portions of overlaps (spatial collisions among nearby boxes), posing additional difficulties for segmentation.
To be effective and keep generality, we choose the fully-parameterized 9-DoF bounding box to handle 6D pose and 3D scale/size estimation. This strategy was studied in marginal space learning for heart chambers detection and aortic valve detection~\cite{zheng2007fast, ghesu2016marginal}, and in incremental parameter learning for ileo-cecal valve detection~\cite{lu2008simultaneous}. However, previous methods employed a hierarchical formulation to estimate 9-DoF parameters in decomposed steps. In contrast, we present a direct and straightforward one-stage 9-DoF bounding box detection strategy.

9-DoF bounding box detection task poses a huge challenge for modern detectors if they are based on heuristic anchor assignment\cite{ren2015faster, liu2016ssd, lin2017focal}, because of the intractable computation expense of intersection-over-union (IoU) in 3D space. It is hard for the widely used post-processing operation of non-maximum suppression (NMS)\cite{girshick2015deformable} on 9-DoF bounding boxes to work efficiently for the same reason. However, in our scenario, due to the uniqueness of each anatomy or query, NMS is not essential. Our Med-Query is built upon DETR, a Transformer-based anchor-free detector~\cite{carion2020end}, to predict the box parameters directly, thus circumvents the otherwise necessary IoU and NMS computations. We further extend another two popular anchor-free detectors of CenterNet~\cite{zhou2019objects} and FCOS~\cite{tian2019fcos} into the 9-DoF setting in our framework to investigate the performance gap between Transformer and convolutional neural network (CNN) architectures. Experimental results show that Transformer-based method empirically exhibits advantages on anatomy parsing problems, which might be attributed to its capability of modeling long-range dependencies and holistic spatial information. 

For a comprehensive anatomy parsing task, only detection and identification is not sufficient and pixel-level segmentation is needed in many scenarios. As we know, the segmentation task should be addressed in high image resolution to reduce imaging resampling artifacts \cite{isensee2021nnu}. However, keeping the whole 3D image in (original) high CT resolution is not computationally efficient. Therefore, detecting to obtain a compact 3D region of interest (ROI) then segmenting to get the instance mask inside ROI becomes the mainstream practice in computationally efficient solutions~\cite{he2017mask, liu2018path, chen2019hybrid, fang2021instances, zheng2007fast, ghesu2016marginal}, which we also follow. 

In this work, we present a unified computing framework for a variety of anatomy parsing problems, with ribs, spine, and abdominal organs as examples. Our method consists of three main stages: task-oriented ROI extraction, anatomy-specific 9-DoF ROI detection, and anatomy instance segmentation. Input 3D CT scans are operated at different resolutions in different processing stages. ROI extractor is trained and tested at an isotropic spacing of 3mm, detector is conducted at 2mm, and the segmentation network is conducted on the cropped sub-volumes, which have a higher image resolution achieved by interpolating input sub-volumes to the desired size and voxel resolution. The whole inference latency is around 3 seconds per CT scan on an NVIDIA V100 GPU, outperforming several highly optimized methods\cite{isensee2021nnu, lenga2018deep} in the rib parsing task. If we query only a subset of all ribs instead, the needed computing time can be further shortened. 

In addition, among these three anatomical structures utilized in this work, the ribs relatively lack a high-quality publicly available dataset. Thus this instance parsing problem has not been as extensively studied as the other two tasks~\cite{yang2021ribseg}. To address this issue, we curate an elaborately annotated instance-level rib parsing dataset, named RibInst, substantially built upon a previously released chest-abdomen CT dataset of rib fracture detection and classification~\cite{ribfrac2020}.  RibInst will be made publicly available to the community by providing a standardized and fair evaluation benchmark for future rib instance segmentation and labeling methods.

Our main contributions are summarized as follows.
\begin{itemize}
    \item We present a Transformer-based Med-Query method for simultaneously estimating and labeling 9-DoF ROI of anatomy in CT scans. To the best of our knowledge, we are the first to estimate the 9-DoF representation in 3D medical imaging using a one-stage Transformer. 

    \item A steerable object/anatomy detection model is achieved by proposing an effective weighted adjacency matrix in the bipartite matching process between query and ground-truth boxes. This steerable attribute enables a novel medical image analysis paradigm that permits to directly retrieve any instance of anatomy of interest and further boost the inference efficiency.

    \item We propose a unified computing framework that can generalize well to a variety of anatomy parsing problems. This framework achieves new state-of-the-art (SOTA) quantitative results on rib parsing task, and performs on par with other SOTA methods on spine parsing and multi-organ segmentation tasks.

    \item Last but not least, we have released publicly a comprehensively annotated instance-level rib parsing dataset of 654 patients, termed RibInst, built on top of a previous chest-abdomen CT dataset~\cite{ribfrac2020}.

\end{itemize}

\section{Related Work}

\noindent {\bf Detection, identification and segmentation.}
Automated parsing techniques have been widely adopted for object/anatomy detection, identification, and segmentation in medical imaging domain~\cite{sharma2010automated, shen2017deep, chen2022recent}. From a specific application perspective, previous work can be roughly categorized into task-specific and universal methods. In rib parsing, existing methods solve rib segmentation and labeling relying heavily on seed points detection and centerline extraction, thus their results can be inaccurate or even fail when seed points are missing or mis-located~\cite{shen2004tracing, klinder2007automated, staal2007automatic, lenga2018deep}. Using auxiliary representations for parsing ribs has also been explored: skeletonized centerlines~\cite{ramakrishnan2011automatic2, wu2012learning} or point cloud~\cite{yang2021ribseg}. 
In spine parsing, some earlier methods adopt complicated heuristic rules to obtain robust labeling results~\cite{yao2006automated, klinder2009automated}. In the era of deep learning, researchers utilize neural networks to perform iterative labeling~\cite{lessmann2019iterative}, conduct detection and segmentation in a multi-stage fashion~\cite{payer2020coarse, sekuboyina2021verse}, capture the vertebrae sequence in a lower-dimensional space~\cite{wang2021automatic, wu2023multi}, and process the spine imaging via Transformers~\cite{tao2022spine, you2023verteformer}. Recent end-to-end segmentation models have demonstrated their versatility and have thus been widely adopted, such as U-Net~\cite{ronneberger2015u, cciccek20163d}, V-Net~\cite{milletari2016v}, and so on. 
Particularly, nnU-Net~\cite{isensee2021nnu} and its variants have achieved cutting-edge performances in various medical imaging segmentation tasks~\cite{antonelli2022medical, ma2021abdomenct, liu2022universal, bian2022artificial}. All existing techniques do not offer steerable 3D object/anatomy detection as query retrieval. In this work, we present a unified computing framework that can generalize well to a variety of anatomy parsing tasks, based on the detection-then-segmentation paradigm in 3D medical image analysis, providing benefits in steerability, robustness, and efficiency.

\noindent {\bf Detection-then-segmentation} paradigm has dominated the instance segmentation problem in photographic images. He et al.\cite{he2017mask} extend Faster R-CNN\cite{ren2015faster} to Mask R-CNN by adding a branch for predicting the segmentation mask of each detected ROI. Several works have emerged from Mask R-CNN, such as PANet \cite{liu2018path}, HTC \cite{chen2019hybrid} and QueryInst \cite{fang2021instances}. There are also other works that do not rely on the explicit detection stage \cite{XinleiChen2019TensorMaskAF, XinlongWang2020SOLOSO, BowenCheng2021PerPixelCI}. The detection-then-segmentation paradigm deserves to be exploited further in 3D medical imaging data where precision and computation efficiency both matter. For rib instance segmentation, previous techniques can not be directly applied since the ribs are orientated obliquely and are so close to neighbouring ones that these regular axis-aligned bounding boxes would be largely overlapped. Partially inspired by \cite{zheng2007fast} and \cite{lu2008simultaneous}, in which 9-DoF parameterized boxes are employed to localize the target region with a hierarchical workflow to estimate the box parameters progressively, we parametrically formulate the detection target for each rib as a 9-DoF bounding box encompassing the rib tightly. This 9-DoF representation can be generalized to other anatomies existing in 3D medical imaging volumes. Direct 9-DoF box estimation on point clouds have been explored recently \cite{wang2019normalized,  lin2021dualposenet, weng2021captra}, demonstrating promising results on this complex pose estimation problem in 3D space. In this work, we tackle a similar task by detecting the 9-DoF bounding box of anatomy in one-stage using 3D CT scans.

\noindent {\bf Anchor-free object detection} techniques have been well studied  \cite{redmon2016you, HeiLaw2020CornerNetDO, XingyiZhou2019BottomUpOD, zhou2019objects, tian2019fcos, ZeYang2019RepPointsPS, carion2020end, XizhouZhu2021DeformableDD, PeizeSun2021SparseRE}.
YOLO\cite{redmon2016you} reports comparable detection performance close to Faster R-CNN\cite{ren2015faster} while running faster. Several anchor-free methods also obtain competitive accuracy with high inference efficiency. Zhou et al.\cite{zhou2019objects} present CenterNet that models an object as the center point of its bounding box so that object detection is simplified as key point estimation and size regression. FCOS\cite{tian2019fcos} takes advantage of all points in a ground-truth box to supplement positive samples and suppresses the low-quality detected boxes with its ``centerness'' branch. Apart from CNN based methods, DETR\cite{carion2020end} comes up with a new representation that views object detection as a direct set prediction problem with Transformer. It learns to produce unique predictions via the bipartite matching loss and utilizes global information with the self-attention mechanism \cite{vaswani2017attention}, beneficial for the detection of large or elongated objects. 
More recently, 3DETR \cite{misra2021end} of a detection model for 3D point clouds has proven that the Transformer-based detection paradigm can be extended to 3D space with minor modifications and achieve convincing results with good scalability in dealing with higher dimensional object detection problems.

\section{Approach}

\subsection{Problem Definition} 
Anatomy parsing focuses on distinguishing each instance in a cluster of similar anatomical structures with a unique anatomical label, e.g., recognizing each rib in a ribcage, splitting each vertebra in a spine, and delineating each organ in the abdominal region in 3D CT scans. Our work distinguishes itself from existing works by its steerable capability that is attributed to a Transformer-based 9-DoF box detection module. We briefly describe our overall detection framework as follows. Given a CT scan containing $N \left(1 \leq N \leq C\right)$ targets, where $C$ is the maximum number of target anatomies in a normal scan (e.g., $C$ is set as 24 for rib parsing in our work according to \cite{glass2002pediatric}), the whole anatomy set can be expressed as $x=\{x_i = (c_i, x_i^p, x_i^s, x_i^a)\mid 1 \leq i \leq N\}$, where $c_i$, $x_i^p$, $x_i^s$, and $x_i^a$ stand for anatomy instance label, center position component, scale component, and angle component of the correspondingly parameterized bounding box, respectively. The goal of the detection task is to find the best prediction set $\hat{x}$ that matches $x$ for each object/organ/anatomy instance. Unless specified otherwise, the term \textit{position} in the following context refers to the position of the box center.

\begin{figure*}[ht]
\centering
\includegraphics[width=\textwidth]{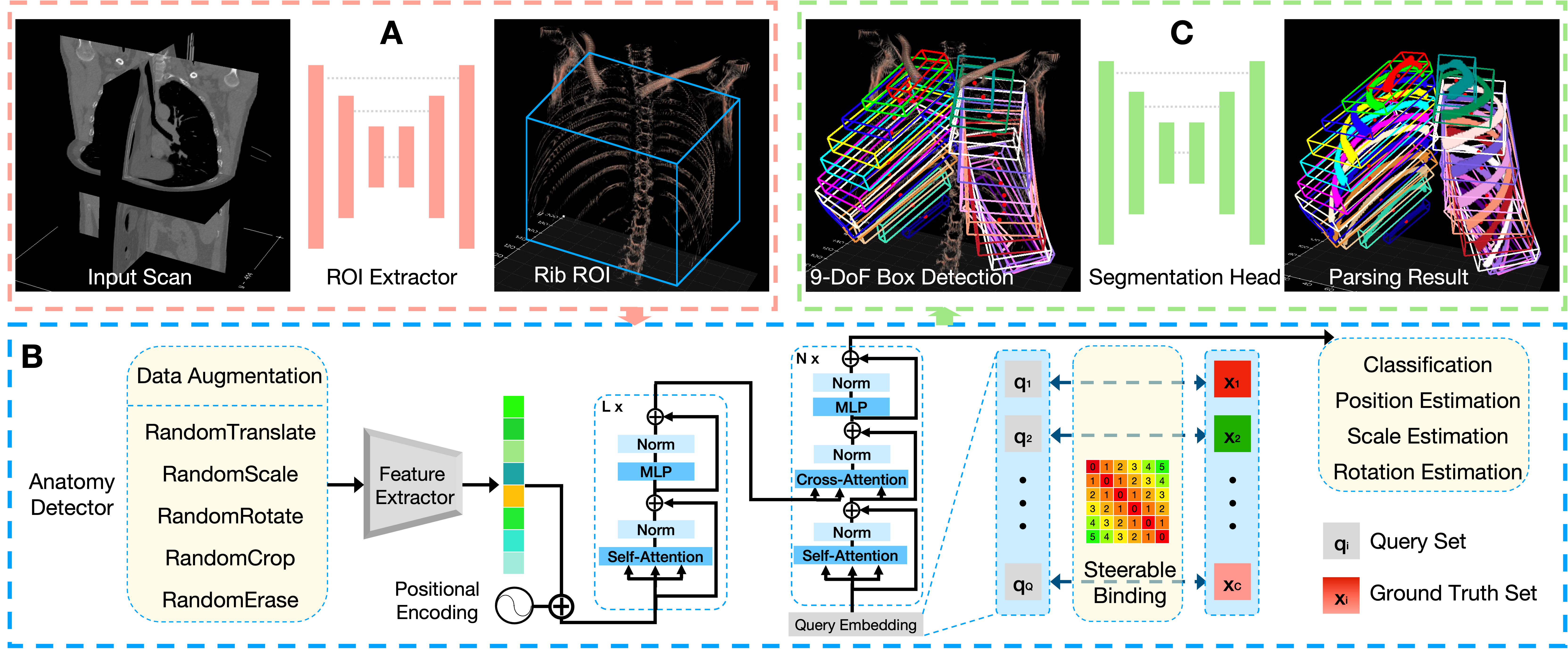}
\caption{An instantiation of Med-Query architecture for rib parsing, which consists of \textbf{A}: a ribcage ROI extractor, \textbf{B}: a steerable 9-DoF parametric rib detector, and \textbf{C}: a stand-alone segmentation head, for robust and efficient rib parsing, i.e., instance segmentation and labeling. In (\textbf{B}) Anatomy Detector, we use an adapted 3D version of ResNet \cite{he2016deep} as feature extractor. The stacked colored illustrative blocks next to the feature extractor represents the flattened spatial features. A set of $q_i$ constitutes the queries, and a set of $x_i$ constitutes the targets.}
\label{architecture}
\end{figure*}

\subsection{9-DoF Box Parameterization}
To obtain a compact parameterization for anatomy localization, we compute the 9-DoF bounding box via principal component analysis (PCA)\cite{jolliffe2016principal}, based on the annotated instance-level 3D segmentation mask. Specifically, three steps are performed: 1) computing the eigenvalues and eigenvectors of covariance matrix based on the voxel coordinates of each anatomy mask; 2) sorting eigenvalues to determine the axes for the local coordinate system with the correspondent eigenvectors; 3) formulating the 9-DoF box in representation of $(x, y, z, w, h, d, \alpha, \beta, \gamma)$ based on the anatomy mask and the local coordinate system. To be more specific, $(x, y, z)$ stands for the box center and is transformed back into the world coordinate system to maintain invariance whenever any data augmentation is applied; $(w, h, d)$ represents the box scale along $x$-axis, $y$-axis, and $z$-axis of the local coordinate system, respectively, and is stored in units of millimeters; $(\alpha, \beta, \gamma)$ stands for the box's Euler angles around $Z$-axis, $Y$-axis, and $X$-axis of the world coordinate system, respectively. An intuitive visual illustration can be found in \cref{concept_illus}. It is noteworthy that this universal parameterization can be customized or simplified, e.g., only keeping the angle of pitch as the rotation parameter may be sufficient to define the vertebra parameterization (\cref{concept_illus}(c)), depending on the specific anatomical characteristics.

\subsection{Med-Query Architecture}
In this section, we describe our full architecture in detail. Based on the steerable anatomy detector, we also enhance our algorithm pipeline with a preceding task-oriented ROI extractor and a subsequent stand-alone segmentation head. The term ``Med-Query'' is not limited to the detector itself but represents the whole processing framework.
For ease of illustration, a schematic flowchart of a concrete application example of rib parsing, is demonstrated in \cref{architecture}.

\noindent {\bf ROI Extractor.} The scales and between-slice spacings of chest-abdomen CT scans vary greatly. To make Med-Query concentrate on the target rib region, a task-oriented ROI extraction which involves only ribcage area would be helpful. Simply using thresholding to get the ROI is not robust, especially for input scans with large FOVs, e.g., from neck to abdomen. To obtain an accurate ROI estimation when handling input CT scans under various situations, we train a simplified U-Net\cite{ronneberger2015u} model to coarsely identify the rib regions in 3D, then the proper ROI box can be inferred by calculating the weighted average coordinates and distribution scope of the predicted rib voxels. At the inference stage, the obtained ROI center is critical and the ROI scale can be adjusted depending on the data augmentation strategies adopted in training. Our ROI extractor is computationally efficient while running on an isotropic spacing of $3mm\times3mm\times3mm$.

\noindent {\bf Steerable Anatomy Detector.} DETR~\cite{carion2020end} revolutionizes object detection by framing it as a direct set prediction problem. It eliminates the need for traditional components like the region proposal network (RPN) and non-maximum suppression (NMS). DETR leverages a Transformer~\cite{vaswani2017attention} architecture for end-to-end object detection, where it predicts objects directly using a bipartite matching loss to align predicted and ground-truth objects. Through parallel decoding of learnable query embeddings, DETR efficiently processes the global context of the input image to produce accurate detection results.
Queries in DETR demonstrate a preference for object spatial locations and sizes in a statistical perspective. If there exists a deterministic binding between the learned query embeddings and anatomical structures, a steerable object parsing paradigm for medical imaging is achievable. 
Medical imaging differs from natural images mainly in two aspects: 1) the semantic targets inside medical imaging scans are relatively more stable with respect to their quantities and absolute/relative positions, even though there exist some local ambiguities; 2) the anatomical label for each instance is unique. These two intrinsic properties constitute the cornerstones of developing a steerable anatomy detector. The remaining obstacle is how to obtain a deterministic/steerable binding.

\begin{figure}[t]
\centering
\includegraphics[width=\linewidth]{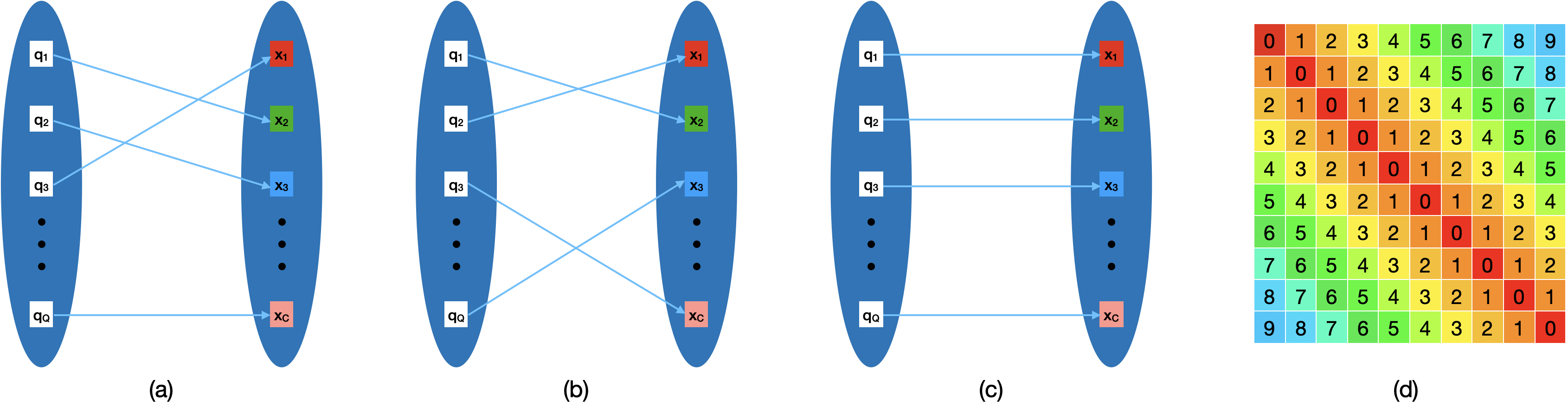}
\caption{The mappings between queries and ground-truth boxes in a transformer-based detector may be random (a)(b). We intend to obtain a fixed binding outcome (c) via a weighted adjacency matrix during training. (d) shows an instantiation of the weighted adjacency matrix with 10 queries (represented by rows) and 10 ground-truth classes (represented by columns).}
\label{binding}
\end{figure}

The original DETR does not have a mechanism to explicitly establish a one-to-one correspondence between queries and ground-truth objects. This can result in varying and inconsistent outcomes after different training attempts, where, for example, the 1st query detects the 2nd class, the 2nd query detects the 3rd class, and the 3rd query detects the 1st class, etc, with this pattern changing each time the model converges. Two simple random binding examples are illustrated in \cref{binding}(a)(b). In order to achieve a steerable system, we expect a fixed binding outcome, as shown in \cref{binding}(c). Formally, we define the query set as $q=\{q_i\mid 0 \leq i \leq Q\}$. Note that in our experiments, the total number of queries is set as $C+1$, which is the number of output channels for the classification branch in each query prediction, with the channel zero as the background class. 

To explicitly guide the bipartite matching process, we propose a weighted adjacency matrix $M \in \mathbb{R}^{(Q+1) \times (Q+1)}$, which can be interpreted as an index cost term to penalize the index mismatch between queries and ground-truth boxes. A matrix instantiation with 10 queries is shown in \cref{binding}(d). As can be seen, the greater the index gap, the higher the index cost. Therefore, assuming a query with index $\sigma(i)$ and its prediction $\hat{x}_{\sigma(i)}$, our matching cost on $\left(\hat{x}_{\sigma(i)}, x_{i}\right)$ can be defined as:
\begin{equation}
\label{match_eq}
\begin{aligned}
\mathcal{C}\left(\hat{x}_{\sigma(i)}, x_{i}\right) = & -\lambda_{c}\hat{p}_{\sigma(i)}\left(c_{i}\right)+ \lambda_{p}\|\hat{x}_{\sigma(i)}^{p} - x_{i}^{p}\|_1\\
&+ \lambda_{s}\|\hat{x}_{\sigma(i)}^{s} - x_{i}^{s}\|_1 + \lambda_{a}\|\hat{x}_{\sigma(i)}^{a} - x_{i}^{a}\|_1 \\
&+\lambda_{m}M[\sigma(i), c_i],
\end{aligned}
\end{equation}
where $c_i$ is the target class label, $\hat{p}_{\sigma(i)}\left(c_{i}\right)$ is the probability prediction for class $c_i$, and $\lambda_c, \lambda_p, \lambda_s, \lambda_a, \lambda_m$ are cost coefficients for classification, position, scale, rotation, and the preset weighted adjacency matrix, respectively. In our implementation, the position values and scale values in the image coordinate system are normalized by the image size $[W, H, D]$, thus they can be merged into the same branch with a \textit{sigmoid} function as its activation. 

The optimal matching $\hat{\sigma}$ is searched using Hungarian algorithm \cite{kuhn1955hungarian}. Then, our loss function is defined similarly as:
\begin{equation}
\label{loss_eq}
\begin{aligned}
\mathcal{L}(\hat{x}, x)=\frac{1}{N}\sum_{i=1}^{N}\left.\Big[-\lambda_{c}\log \hat{p}_{\hat{\sigma}(i)}\left(c_{i}\right)+ \lambda_{p}\|\hat{x}_{\hat{\sigma}(i)}^{p} - x_{i}^{p}\|_1 \right. \\ 
\phantom{=\;\;}\left. + \lambda_{s}\|\hat{x}_{\hat{\sigma}(i)}^{s} - x_{i}^{s}\|_1
+ \lambda_{a}\|\hat{x}_{\hat{\sigma}(i)}^{a} - x_{i}^{a}\|_1 \right.\Big],
\end{aligned}
\end{equation}
where the coefficients are kept consistent with those in \cref{match_eq}. For queries that are matched to the background class, only the loss of classification is taken into account.
Note that we merely use L1 Loss for the geometrical 9-DoF box regression. No IoU-related loss \cite{rezatofighi2019generalized} is used in bipartite matching or loss computing, which is different from \cite{carion2020end} and \cite{misra2021end}.
The carefully designed weighting matrix only takes effects at the matching stage and it will gradually guide the bipartite matching process between queries and ground-truth boxes to achieve a stable binding.

\noindent {\bf Segmentation Head.} 
The input data for the aforementioned detector is isotropic with a spacing of $2mm\times2mm\times2mm$ per voxel which is finer than the ROI extraction step. To obtain high accuracy instance segmentation results with affordable training expense, we adopt a stand-alone U-Net\cite{ronneberger2015u} to segment each anatomy independently with a finer spatial resolution in a locally cropped FOV. Specifically, each detected 9-DoF bounding box from our steerable anatomy detector is slightly expanded to ensure that all the anatomical structures of interest are included, and then used to crop a sub-volume from the input CT volume. This sub-volume is interpolated as needed to achieve the desired input size for the segmentation head, which may vary across different organs. Then, the segmentation head performs a binary segmentation for all sub-volumes, with the anatomy of interest being segmented as foreground and other tissues (including neighboring anatomies of the same category) as background. After this, all predicted binary masks are merged back with their corresponding labels and spatial locations into the original CT scan coordinate system to form the final instance segmentation output. Our detector and segmentation head is designed to be disentangled, so there are good flexibility and scalability to employ even more powerful segmentation networks than U-Net~\cite{ronneberger2015u}. Furthermore, benefiting from the steerable nature of the proposed detector, the segmentation head can be invoked dynamically per detection box at request, to save computational cost if necessary.

\subsection{Data Augmentation}\label{sec:data_aug}
For better model generalization and training efficiency, we employ both online and offline data augmentation schemes.

\noindent {\bf Offline Scheme.} We perform \textit{RandomCrop} along the $Z$-axis to imitate (largely) varying data input FOVs in realistic clinical situations, where FOVs can contain or cover the targeted object/anatomy region completely or only partially. Spatial cropping may truncate some obliquely oriented anatomies, whose 9-DoF parameterizations need to be recomputed. Online computation is time-consuming so we conduct this operation offline.

\noindent {\bf Online Schemes.} Since the anatomy position, orientation, and scale vary case by case, local spatial ambiguities exist among different CT scans, posing great challenges for identifying each anatomy correctly and precisely. We perform 3D \textit{RandomTranslate}, \textit{RandomScale}, and \textit{RandomRotate} to add more input variations and alleviate this problem. These operations can be conducted efficiently without 9-DoF box recomputation, so they are performed online during the training process. According to \cite{glass2002pediatric}, 5\%-8\% of normal individuals may have 11 pairs of ribs. This raises an obstacle because we have insufficient training data exhibiting this pattern of anatomy variation. Hence, for the rib parsing task, we particularly use \textit{RandomErase} to remove the bottom pair of ribs in the training set with a certain probability ratio, to alleviate the possible over-prediction problem. No additional data augmentation is used otherwise.

\section{Experiments}

\subsection{Datasets}
\noindent {\bf RibInst} dataset is constructed from a public CT dataset that was used as the rib fracture evaluation benchmark in \textit{RibFrac} challenge\cite{ribfrac2020}. This original dataset consists of 660 chest-abdomen CT scans with 420 for training, 80 for validation and 160 for testing. We comply with the dataset splitting rule and annotate each rib within each CT scan with spatially ordered and unique labels. There are 6 cases with extremely incomplete FOV in the validation set, making it infeasible to identify the correct rib labels. Therefore RibInst dataset has the remaining 654 CT scans. Note that a dataset with 490 CT scans annotated with binary rib segmentation masks and labeled rib centerlines was released \cite{yang2021ribseg}. Compared with it, RibInst is more concise and comprehensive, for conducting rib instance segmentation evaluation without centerline extraction.

\noindent {\bf CTSpine1K} provides 1,005 CT scans with instance-level vertebra annotation for spine analysis ~\cite{deng2021ctspine1k, liu2022universal}. The dataset is curated from four previously released datasets ranging from head-and-neck imaging to colonography scans.
It serves as a benchmark for spine-related image analysis tasks, such as vertebrae segmentation and labeling. There are 610 scans for training and the remaining are for validation and testing.

\noindent {\bf FLARE22} stands for Fast and Low-resource semi-supervised Abdominal oRgan sEgmentation in CT, a challenge held in MICCAI 2022. The dataset is curated from three abdomen CT datasets. 
The segmentation targets contain 13 organs: liver, spleen, pancreas, right kidney, left kidney, stomach, gallbladder, esophagus, aorta, inferior vena cava, right adrenal gland, left adrenal gland, and duodenum.
As a semi-supervised task, the training set includes 50 labeled CT scans with pancreas disease and 2,000 unlabeled CT scans with liver, kidney, spleen, or pancreas diseases. 
There are also 50 CT scans in the validation set and 200 CT scans in the test set with various diseases. For more information, readers are referred to the challenge website\footnote{https://flare22.grand-challenge.org}.

\subsection{RibInst Curation Details}
{\noindent {\bf Iterative Annotation.} First of all, we need to obtain well-annotated binary rib segmentation mask for each volume even if manually annotating voxels in 3D space is a time-consuming, labor-intensive task. We adopted an automatic tool to improve annotation efficiency, which has an operating pipeline of 1) separating bones using thresholding and morphology operations, 2) removing the spine and the sternum, and 3) distinguishing rib from other remaining bones with a pre-trained machine learning model. Using this tool, we collected 50 coarse masks with hollow components, thus binary closing operation was performed on each mask to include rib marrow regions. Two radiologists with over 5 years of experience manually removed the cartilago costalis and refined the boundaries between the ribs and spine for each scan. Then, we trained one U-Net \cite{ronneberger2015u} and predicted foregrounds (rib masks) for another 50 cases among unannotated scans using the trained model, followed by a new round of manual refinement. After several rounds of this iterative annotation process, we obtained high-quality binary segmentation for all ribs in this dataset.

\noindent {\bf Instance Labeling.} Given the binary segmentation, we wish to efficiently set a unique label per rib in each CT scan. An interactive annotation tool was developed to fulfill this goal. The label for each rib within one connected component is determined by one click and the subsequently entered number. The maximal number was set to 24, as 12 pairs of ribs may present at most common scans, 5\%-8\% of normal individuals may have 11 pairs, while supernumerary ribs may rarely be seen as normal variant according to \cite{glass2002pediatric}.
The instance label for each rib increases in the order of top to bottom and right to left in the patient view.

\noindent {\bf Quality Control.} Despite its high efficiency of our instance labeling tool, it is hard (if not impossible) to avoid mistakes when observing and labeling in 3D space, especially for identifying elongated and oblique objects such as ribs. Thus it is essential to perform quality control to improve and assure the overall annotation quality. All labeled masks were projected into 2D space via cylindrical projection~\cite{tobon2017unfolded} to obtain visually intuitive demonstrations. Rib labels were mapped to common strings for the sake of readability and attached to the corresponding ribs. Mislabeled ribs or bad segmented components were easy to be distinguished in this 2D image. When mistakes were found, we turned back to previous stages to manually repair the segmentation masks or correct the labels.

\begin{table*}[!t]
    \centering
    \caption{\upshape Detection and identification results on the test set of RibInst. *Note that CenterNet and FCOS demonstrated here are significantly developed and modified to fit into our task and equipped with the proposed 9-DoF box representation. GFLOPs are computed under the same input size $256\times256\times256$. The $p$-values are calculated using the Id.Rate of Med-Query (Ours) as a reference to determine the difference between the baseline methods and our method, using a paired $t$-test.}
    \label{tbl:quant_det_rib}
    \setlength{\tabcolsep}{0.5mm}{
    \begin{tabular}{lcccccccc}
    \toprule[1pt]
    Methods & \#params(M) & GFLOPs$\downarrow$ & Id.Rate(\%)$\uparrow$ & $p$ & $\operatorname{P_{mean}}(\text{mm})$$\downarrow$  & $\operatorname{S_{mean}}(\text{mm})$$\downarrow$ & $\operatorname{A_{mean}}(^{\circ})$$\downarrow$ & Latency(s)$\downarrow$\\
    \midrule
    CenterNet*~\cite{zhou2019objects}  & 1.3 & 1897.7  & 94.9$\pm$9.4    &  2e-7    &\textBF{3.032$\pm$0.662}  &\textBF{2.333$\pm$0.659}  & 2.890$\pm$0.651        & 2.575$\pm$0.136      \\
    FCOS*~\cite{tian2019fcos}       & 19.4 & 1848.5 & 94.1$\pm$7.7    &  6e-8   & 3.100$\pm$0.792         & 2.516$\pm$0.708         &\textBF{1.786$\pm$0.355} & 0.745$\pm$0.025   \\
    Med-Query  & 9.6 & 90.7 &\textBF{97.0$\pm$4.2} & - & 4.702$\pm$2.107         & 2.784$\pm$0.952         & 2.185$\pm$0.414        &\textBF{0.065$\pm$0.008}\\
    \bottomrule[1pt]
    \end{tabular}}
\end{table*}

\subsection{Training Details}
Our implementation uses PyTorch \cite{paszke2019pytorch} and is partially built upon MMDetection\cite{chen2019mmdetection}. The detector in Med-Query is trained with AdamW optimizer \cite{loshchilov2017decoupled} in the initial setting of $lr=4\mathrm{e}{-4}, \beta_1=0.9, \beta_2=0.999, weight\_{decay}=0.1$. The total training process contains 1000 epochs: the first 200 epochs to linearly warm up and the learning rate $lr$ is reduced by a factor of 10 at the 800th epoch. It costs about 20 hours on 8 V100 GPUs with a single GPU batch size of 8. We empirically set coefficients $\lambda_c=1, \lambda_p=10, \lambda_s=10, \lambda_a=10, \lambda_m=4$. The network outputs of the fourth stage (C4) from an adapted 3D ResNet50 are used as the spatial features for Transformer encoding. The Transformer component consists of a one-layer encoder and a three-layer decoder, which we have empirically found to be sufficient for our task. We follow \cite{misra2021end} to set a dropout of 0.1 in encoder and 0.3 in decoder. The \texttt{MultiHeadAttention} block in both the encoder and decoder has 4 heads. The positional encoding length is set to 128 per axis, resulting in a total embedding dimensions of 384. We use data augmentations mentioned in \cref{sec:data_aug} to generate diverse training data both offline and on-the-fly. The maximum translation distance from the reference center (i.e., anatomy mask center in our experiments) is 20mm, the scaling range is [0.9, 1.1], and the rotation range is [-15$^\circ$, 15$^\circ$]. The ROI extractor and segmentation head are trained separately, for the sake of training efficiency.
All models in our experiments, including the feature extractor of the anatomy detector, are trained from scratch.

\subsection{Performance Metrics}
\noindent {\bf Detection and Identification.}
To evaluate the 9-DoF box detection and identification performance, we refer to the practice of \textit{VerSe} \cite{sekuboyina2021verse}, a vertebrae labeling and segmentation benchmark. We extend the Identification Rate (Id.Rate) computation in \textit{VerSe} by considering all factors of the label accuracy, center position deviation ($\operatorname{P_{mean}}$), scale deviation ($\operatorname{S_{mean}}$) and angle deviation ($\operatorname{A_{mean}}$), to evaluate 9-DoF box predictions from different methods. To be more specific: given a CT scan containing $N$ ground-truth anatomy boxes and the true location of the $i^{th}$ anatomy box denoted as $x_i$ with its predicted box denoted as $\hat{x}_{i}$ (with $M$ predictions in total), the anatomy $i$ is correctly identified if $\hat{x}_{i}$ among $\{\hat{x}_{j} \forall j \in \{1, 2, \ldots, M\}\}$ is the closest box predicted to $x_i$ and subjects to the following conditions:
\begin{equation}
\label{det_conditions}
\left\{
\begin{array}{r}
     \|\hat{x}_{i}^{p} - x_{i}^{p}\|_2 < 20\operatorname{mm}, \\
     (\sum_{k=1}^{3}\|\hat{x}_{i}^{s_k} - x_{i}^{s_k}\|_1)/3 < 20\operatorname{mm}, \\
     (\sum_{k=1}^{3}\|\hat{x}_{i}^{a_k} - x_{i}^{a_k}\|_1)/3 < 10^{\circ},
\end{array}
\right.
\end{equation}
where the superscript $p$ denotes the center position component of the box, the superscript $s_k$ denotes the scale component in axis $k$, and the superscript $a_k$ represents the Euler angle component rotated around axis $k$. Given a CT scan, if there are $R$ anatomies correctly identified, then Id.Rate is defined as $\operatorname{Id.Rate} = \frac{R}{N}$. The center position deviation is computed as $\operatorname{P_{mean}} = (\sum_{i=1}^{R}\|\hat{x}_{i}^{p} - x_{i}^{p}\|_2)/R$. Similarly, the scale deviation is computed as $\operatorname{S_{mean}} = (\sum_{i=1}^{R}\sum_{k=1}^{3}\|\hat{x}_{i}^{s_k} - x_{i}^{s_k}\|_1)/3R$, and the angle deviation is computed as $\operatorname{A_{mean}} = (\sum_{i=1}^{R}\sum_{k=1}^{3}\|\hat{x}_{i}^{a_k} - x_{i}^{a_k}\|_1)/3R$. Note that we only compute the average deviations of the identified anatomies, as unpredicted anatomies or mislocated predictions have been reflected by the Id.Rate index. 

\noindent {\bf Segmentation.}
We employ the widely-used Dice similarity coefficient (DSC), 95\% Hausdorff distance (HD95) and average symmetric surface distance (ASSD) as segmentation metrics. 
For missing anatomies, HD and ASSD are not defined. 
We follow the practice of \cite{sekuboyina2021verse} to ignore such anatomies when computing the averages. Those missing anatomies will be reflected on DSC and Id.Rate. We utilize a publicly available evaluation toolkit\footnote{https://github.com/deepmind/surface-distance} to compute the surface measures.

\begin{figure}[t]
\centering
\includegraphics[width=\linewidth]{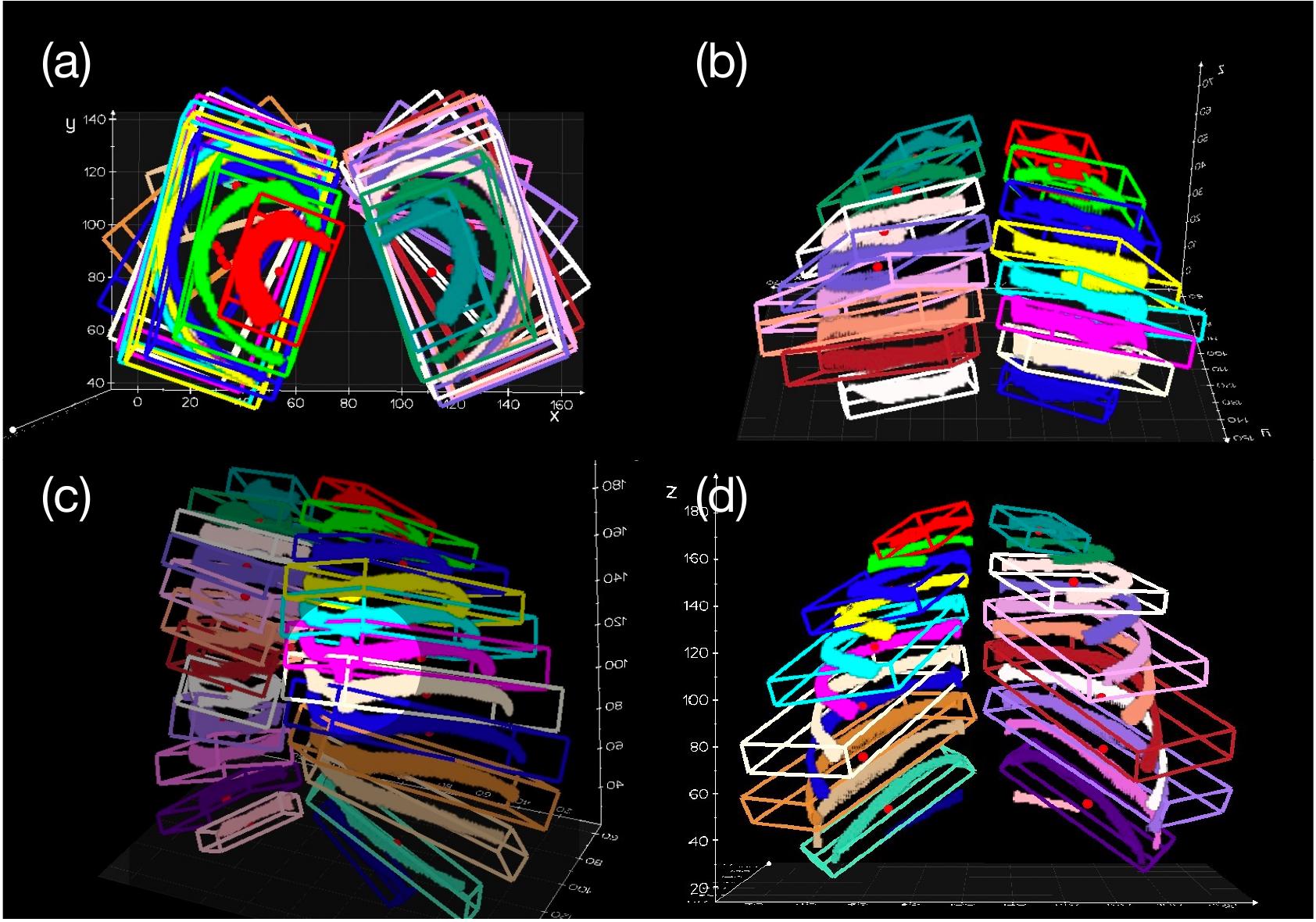}
\caption{Detection visualizations show that our 9-DoF predictions enclose the ground-truth rib masks accurately. (a) Normal results in superior-to-inferior view. (b) A limited FOV case in posterior-to-anterior view. (c) A case with rib adhesions. (d) Only odd-number labels are queried. Ground-truth masks are rendered as visual reference.}
\label{fig:det_visual}
\end{figure}

\subsection{Main Results}
We report comprehensive evaluation results on RibInst from rib instance detection and segmentation. Besides, we validate the performance generality of our framework on CTSpine1K and FLARE22 datasets, and report the segmentation performances compared to several strong baseline methods. We calculate $p$-values using the results of our method as a reference to assess the statistical significance of the differences between the baseline methods and our method, employing a paired $t$-test. $p < 0.05$ indicates a statistical significant improvement.

\begin{table}
\centering
    \caption{\upshape Segmentation results on the test set of RibInst. $\dag$A centerline-based method proposed in~\cite{lenga2018deep}. *Adapted versions, also equipped with ROI extractor and segmentation head. ``nnU-Net cas.'' means coarse-to-fine cascade version and ``nnU-Net'' means 3D full resolution version. Abbreviations: ``TotalSeg''-TotalSegmentator, ``DSC''-Dice Similarity Coefficient, ``HD95''-95\% Hausdorff Distance, ``ASSD''-Average
Symmetric Surface Distance.}
    \label{tbl:quant_seg_rib}
    \setlength{\tabcolsep}{0.78mm}{
        \begin{tabular}{l ccccc}
        \toprule[1pt]
        Methods           & DSC(\%)$\uparrow$ & $p$ & HD95(mm)$\downarrow$  & ASSD(mm)$\downarrow$ & Latency(s)$\downarrow$\\
        \midrule
        TotalSeg & 77.1$\pm$7.1 & 1e-58& 15.76$\pm$3.145 & 2.088$\pm$3.959 & - \\
        Centerline\dag    &  84.6$\pm$21.2 & 4e-4 & 13.84$\pm$32.86    & 3.437$\pm$10.52         & 18.50$\pm$3.182     \\
        CenterNet*        &  89.5$\pm$8.8   & 0.008 & 1.826$\pm$3.031    & 0.725$\pm$1.821         & 4.793$\pm$1.259     \\
        FCOS*             &  88.1$\pm$7.0  & 3e-5 & \textBF{1.137$\pm$1.409}  & \textBF{0.352$\pm$0.947} & 3.241$\pm$1.026     \\
        nnU-Net cas.      &  89.3$\pm$3.5  & 0.002 & 2.421$\pm$4.117    & 0.497$\pm$0.899         & 183.0 $\pm$ (-)     \\ 
        nnU-Net           &  89.7$\pm$7.8  & 0.024 & 4.498$\pm$5.524    & 0.900$\pm$1.658         & 252.0 $\pm$ (-)     \\
        Med-Query         & \textBF{90.9$\pm$7.4} & - &1.644$\pm$3.080 & 0.438$\pm$1.486 &\textBF{2.591$\pm$0.977}\\
        \bottomrule[1pt]
        \end{tabular}}
\end{table}

\begin{table*}[t]
	\centering
	\caption{\upshape Organ-specific DSC (\%) on the validation set of FLARE22. Abbreviations: ``TotalSeg''-TotalSegmentator, ``Liv.''-Liver, ``RKid''-Right Kidney, ``Spl.''-Spleen, ``Pan.''-Pancreas, ``Aor.''-Aorta, ``IVC''-Inferior Vena Cava, ``RAG''-Right Adrenal Gland, ``LAG''-Left Adrenal Gland, ``Gal.''-Gallbladder, ``Eso.''-Esophagus, ``Sto.''-Stomach, ``Duo.''-Duodenum, ``LKid''-Left Kidney, ``ens.''-ensemble, ``pre.''-pre-training, ``mDSC''-mean DSC.}
	\label{tbl:quant_flare}
	\resizebox{\textwidth}{!}{
	\begin{tabular}{lcccccccccccccc}
		\toprule[1pt]
		Methods & Liv. & RKid & Spl. & Pan. & Aor. & IVC & RAG & LAG & Gal. & Eso. & Sto. & Duo. & LKid & mDSC$\uparrow$\\
		\midrule
            TotalSeg~\cite{wasserthal2023totalsegmentator} & 96.7 & 87.5 & 96.9 & 86.5 & 90.4 & 89.9 & 77.1 & 75.3 & 82.5 & 83.3 & 90.9 & 70.4 & 90.0 & 85.9 \\
		nnU-Net~\cite{isensee2021nnu} & 97.7 & 94.1 & 95.8 & 87.2 & 96.8 & 87.8 & 83.0 & 80.1 & 76.5 & 89.2 & 89.9 & 77.1 & 91.1 & 88.2\\
		nnU-Net ens. & 97.9 & \textbf{94.8} & 96.0 & 88.6 & \textbf{96.9} & 89.7 & \textbf{83.8} & \textbf{81.9} & 78.7 & \textbf{90.1} & 90.7 & \textbf{79.2} & 92.0 & 89.2\\
		\midrule
		Swin UNETR~\cite{tang2022self} & 96.5 & 91.2 & 94.2 & 84.6 & 93.0 & 86.5 & 75.8 & 74.2 & 77.1 & 79.0 & 88.6 & 76.5 & 88.7 & 85.0\\
		Swin UNETR pre. & 96.4 & 92.1 & 95.2 & 88.1 & 93.7 & 86.2 & 79.4 & 79.1 & 79.2 & 81.8 & 89.5 & 79.0 & 87.9 & 86.7\\
		Med-Query & \textbf{98.0} & 94.5 & \textbf{97.2} & \textbf{89.0} & 96.6 & \textbf{90.3} & 82.4 & 80.6 & \textbf{86.1} & 87.4 & \textbf{91.5} & 78.7 & \textbf{93.7} & \textbf{89.7}\\
		\bottomrule[1pt]
	\end{tabular}}
        \vspace{-1em}
\end{table*}

\begin{figure*}[h]
\centering
\includegraphics[width=\textwidth]{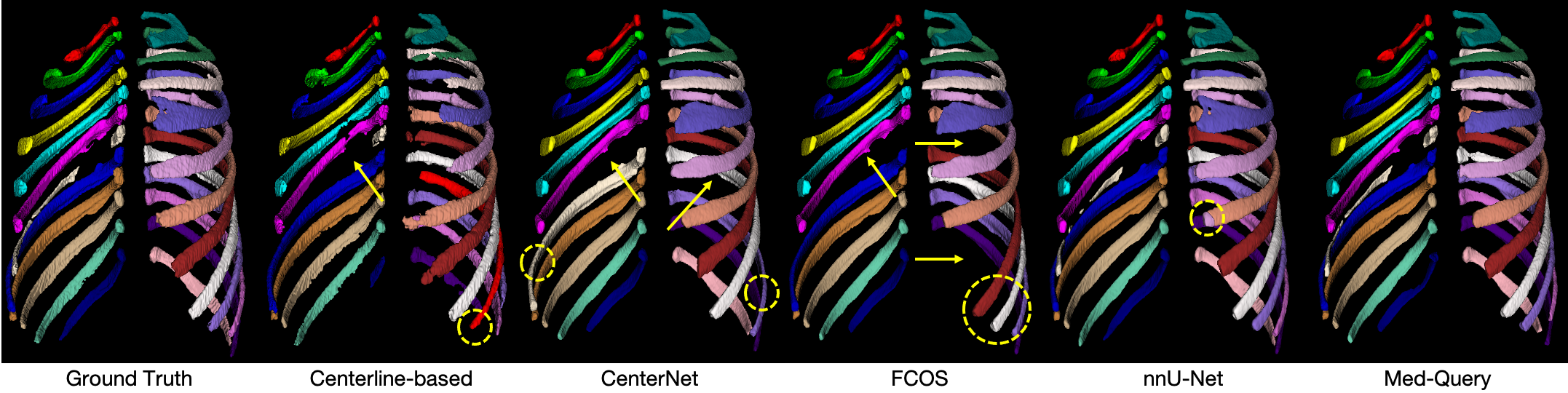}
\caption{An example with broken structures in RibInst. Missing or wrong labels are marked using golden arrows and dashed circles, respectively.}
\label{fig:seg_visual}
\end{figure*}

\noindent {\bf Detection Evaluation on RibInst.} To the best of our knowledge, there are no existing 6-DoF baselines specifically tailored for the rib parsing task for comparison. Consequently, we constructed two strong baselines, FCOS~\cite{tian2019fcos} and CenterNet~\cite{zhou2019objects}, and enhanced them with our 9-DoF capabilities. The results are summarized in~\cref{tbl:quant_det_rib}. As observed, Med-Query achieves the best identification rate of 97.0\% with a moderate amount of parameters. We also compute $p$-values on Id.Rate results between different methods, taking Med-Query as the reference. The $p$-values of CenterNet and FCOS are 2e-7 and 6e-8, respectively, indicating a statistical significant improvement of our method.
The Transformer's capability of capturing long-range dependencies and holistic information appears to be suitable for solving this sequence modeling problem.
Notably, compared to pure CNN architectures, Med-Query can infer at least 10$\times$ faster in terms of latency. This may benefit from not relying on additional upsampling layers in \cite{zhou2019objects} or dense prediction heads in \cite{tian2019fcos}. Additionally, we have observed some interesting differences between Med-Query and traditional object detectors \cite{zhou2019objects,tian2019fcos} that recognize objects individually while ignoring relations between objects\cite{hu2018relation}. In our case, the main reason why FCOS and CenterNet could fail is that they sometimes assign different labels to the same rib or same label to different ribs, subsequently leading to missing ribs as shown in \cref{fig:seg_visual}. These traditional detectors do not explicitly consider the anatomy uniqueness and spatial relationship/dependency between rib labels. Med-Query behaves in another way that solves a label assignment problem rather than an unconstrained object detection task, so that the relations between targets are explicitly constructed and enforced through the steerable label assignment strategy and the intrinsic self-attention mechanism in Transformer models \cite{vaswani2017attention}. On the other hand, Med-Query is slightly inferior in local regression indicators as shown in \cref{tbl:quant_det_rib}. Med-Query can correctly identify more ribs with global information, but including those challenging ribs in the calculation of deviations may deteriorate the quantitative performances on localization errors. It is noteworthy that all these optimized models achieved small regression deviations with respect to $\operatorname{P_{mean}}$, $\operatorname{S_{mean}}$, and $\operatorname{A_{mean}}$. It should be noted that we have improved FCOS and CenterNet with our one-stage 9-DoF detection strategy. The above results demonstrate that the challenging 9-DoF parameter estimation problem can be solved effectively. Some qualitative results of Med-Query in rib detection and labeling are shown in \cref{fig:det_visual}. Our 3D detection visualization tool is developed based on \textit{vedo} \cite{musy2021vedo}.

\begin{figure*}[t]
\centering
\includegraphics[width=\textwidth]{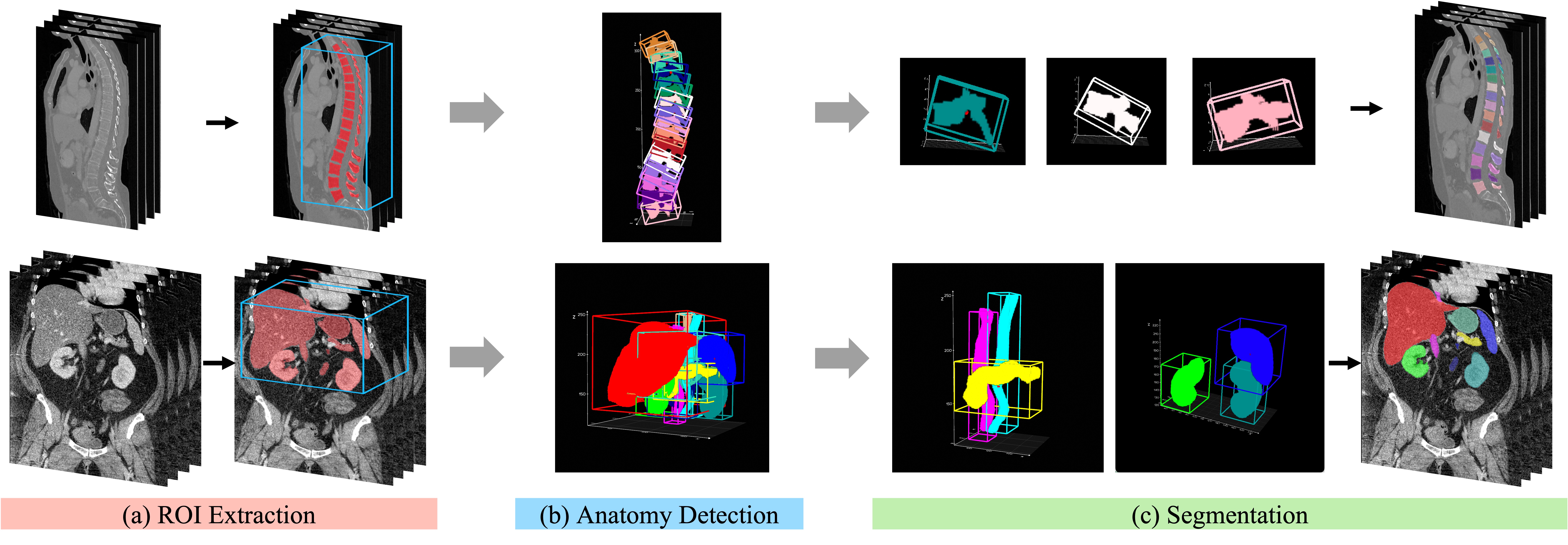}
\caption{Med-Query workflow on spine parsing and abdominal multi-organ segmentation tasks. Similar to rib parsing, this process also consists of three main steps: (a) ROI extraction stage, (b) anatomy detection stage, and (c) instance segmentation and merging stage. Notably, on spine parsing, we only keep the angle of pitch as the rotation parameter to define the vertebra parameterization, and on multi-organ segmentation, a simpler 6-DoF representation (three angles are set to 0) is more appropriate.}
\label{fig:vert_organ_vis}
\end{figure*}

\noindent {\bf Segmentation Evaluation on RibInst.}
As shown in \cref{tbl:quant_seg_rib}, the widely-used self-configuring nnU-Net~\cite{isensee2021nnu} achieves a good DSC of 89.7\% with a long latency of 252 seconds. Its coarse-to-fine cascade version also achieves a competitive DSC of 89.3\%. nnU-Net is trained with sampled patches, therefore the lack of global information leads to noticeable label confusion, significantly revealed by HD95 of 4.498 mm.
We also reimplement a centerline-based method~\cite{lenga2018deep} to investigate how the parsing results relies heavily on centerline extraction and heuristic rules. Additionally, we also make a comparison to the results of TotalSegmentator~\cite{wasserthal2023totalsegmentator}, a model that can segment over 100 anatomical structures. However, TotalSegmentator does not achieve very competitive results on our RibInst dataset, with average scores of 77.1\% for DSC, 15.76 mm for HD95, and 2.088 mm for ASSD, respectively. The primary reason is that the TotalSegmentator training dataset has incomplete annotations for ribs, not covering the areas adjacent to the spine, leading to the trained model generating numerous false negatives in these regions.
Among all listed methods in \cref{tbl:quant_seg_rib}, Med-Query achieves the best DSC of 90.9\%.
The $p$-values reveal the improvement on segmentation is also significant.
Notably, the whole pipeline of Med-Query can be finished in 2.591 seconds. The efficiency is attributed to our detection-then-segmentation paradigm which detects on low resolution with global FOV and segments on high resolution with local FOV. 
A qualitative comparison on a representative patient case with rib fractures is shown in \cref{fig:seg_visual}, demonstrating the robustness of Med-Query.

\noindent {\bf Segmentation Evaluation on CTSpine1K.}
To be aligned with the evaluation protocol in ~\cite{liu2022universal}, we report segmentation metrics on vertebra-level average scores computed on all scans in the validation and test sets (HD is employed instead of HD95). As in \cref{tbl:quant_ctspine}, we compare Med-Query with several strong baselines, including nnFormer~\cite{zhou2021nnformer}, nnU-Net~\cite{isensee2021nnu}, CPTM~\cite{liu2022universal}, Payer et al.~\cite{payer2020coarse}, and TotalSegmentator~\cite{wasserthal2023totalsegmentator}. In this comparison, the generalist TotalSegmentator and Payer et el., the top-ranked specialist method in \textit{VerSe} challenge~\cite{sekuboyina2021verse}, achieve the best volume and surface metrics, respectively. Med-Query performs on par with these state-of-the-art methods, demonstrating its robustness and applicability.

\begin{table*}[!t]
\begin{minipage}{0.42\textwidth}
    \centering
    \caption{\upshape Segmentation results on the validation set and test set of CTSpine1K. The results of nnFormer, nnU-Net, and CPTM are quoted from CPTM paper. Abbreviations: ``DSC''-Dice Similarity Coefficient, ``HD''-Hausdorff Distance, ``ASSD''-Average Symmetric Surface Distance.}
        \setlength{\tabcolsep}{1mm}{
        \begin{tabular}{l ccc}
        \toprule[1pt]
        Methods           & DSC(\%)$\uparrow$ & HD(mm)$\downarrow$  & ASSD(mm)$\downarrow$\\
        \midrule
        nnFormer~\cite{zhou2021nnformer}          &  74.3   & 11.56  &  -     \\
        nnU-Net~\cite{isensee2021nnu}           & 84.2   & 9.02   &  -     \\
        CPTM~\cite{liu2022universal}              &  84.5   & 9.03   &  -     \\
        Payer et al.~\cite{payer2020coarse}      &  81.9   & \textBF{7.07}   &  \textBF{0.78}  \\
        TotalSeg~\cite{wasserthal2023totalsegmentator}     & \textBF{85.4}   & 8.28    & 0.99 \\
        Med-Query         & 85.0  & 8.68    & 1.16    \\
        \bottomrule[1pt]
        \end{tabular}}
    \label{tbl:quant_ctspine}
\end{minipage}
\begin{minipage}{0.58\textwidth}
\caption{\upshape Id.Rate evaluations in different operations and different measurement thresholding combinations. \textit{RandomTranslate}, \textit{RandomScale}, and \textit{RandomRotate} are performed jointly, thus are abbreviated as ``T.S.R''. ``Re.Dist.'' represents relative distance constraints. ``PE'' is short for positional encoding. The values in the top row of the right column represent the different maximum tolerances used for translation, scale and rotation in \cref{det_conditions}.}
\label{tbl:ablation_aug}
\setlength{\tabcolsep}{0.8mm}{
\begin{tabular}{cccccc|ccc}
\toprule[1pt]
\multicolumn{1}{c}{\multirow{2}{*}{No Aug.}} & \multicolumn{3}{c}{Data Aug.} & \multirow{2}{*}{Re.Dist.} & \multirow{2}{*}{w/o PE} & \multirow{1}{*}{20mm20mm} & \multirow{1}{*}{10mm10mm} & \multirow{1}{*}{10mm10mm}\\ \cmidrule{2-4}
\multicolumn{1}{c}{} & \multicolumn{1}{c}{T.S.R} & \multicolumn{1}{c}{Crop} & \multicolumn{1}{c}{Erase} & \multicolumn{1}{c}{} & \multicolumn{1}{c|}{} & \multirow{1}{*}{10$^{\circ}$} & \multirow{1}{*}{10$^{\circ}$} & \multirow{1}{*}{5$^{\circ}$} \\ 
\hline
\checkmark &            &            &            &            &             &   83.1$\pm$22.1 & 49.0$\pm$29.6 & 33.7$\pm$25.9 \\
           & \checkmark &            &            &            &             &   94.4$\pm$11.3 & 84.8$\pm$20.8 & 79.2$\pm$21.8 \\
           & \checkmark & \checkmark &            &            &             &   96.2$\pm$4.1  & 89.0$\pm$16.7 & 84.2$\pm$18.7 \\
           & \checkmark & \checkmark & \checkmark &            &             &   96.5$\pm$5.3  & 90.7$\pm$17.0 & 87.0$\pm$17.7 \\
           & \checkmark & \checkmark & \checkmark & \checkmark &             &   \textbf{97.0$\pm$4.2} & 91.8$\pm$15.0 & 87.4$\pm$17.6 \\
           & \checkmark & \checkmark & \checkmark & \checkmark & \checkmark  &   96.2$\pm$5.9 & \textbf{92.0$\pm$14.9} & \textbf{87.9$\pm$15.8} \\
\toprule[1pt]
\end{tabular}}
\end{minipage}
\end{table*}

\begin{figure*}[t]
\centering
\includegraphics[width=\textwidth]{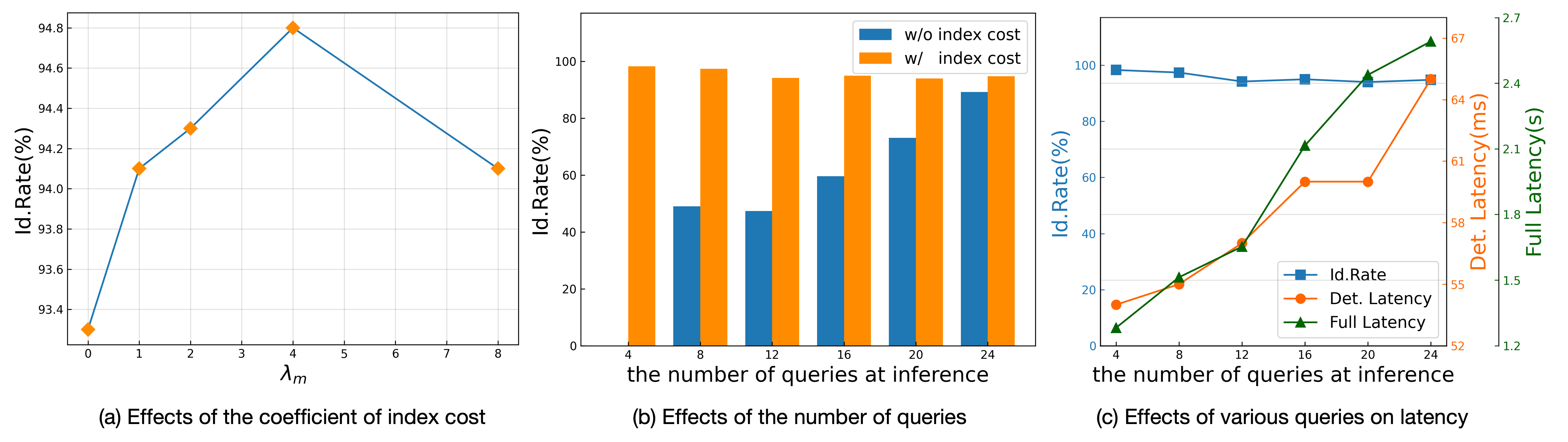}
\caption{Ablation studies on the coefficient of the proposed index cost and the number of query embeddings at inference. (a) The coefficient $\lambda_m = 4$ yields the optimal Id.Rate on rib labeling task. (b) Med-Query with index cost exhibits stable
performance when the number of queries varies. (c) The latency of Med-Query can be further reduced with less queries required.}
\label{fig:ablation}
\end{figure*}

\noindent {\bf Segmentation Evaluation on FLARE22.}
We report the evaluation results on FLARE22 validation leaderboard that gives the detailed performance evaluation at organ-level. In~\cref{tbl:quant_flare}, nnU-Net and its ensemble version are merely trained on labeled data, as shown, they can achieve very good results on the validation set even with only 50 labeled data. The ensembled nnU-Net (assembly of 12 models from different settings and different folds in cross-validation) can obtain a 1\% improvement on mean DSC. Considering the semi-supervised setting of this task, we use nnU-Net as the teacher model to generate pseudo labels for the unlabeled part of the training set, as a straightforward but effective semi-supervised strategy. 
Based on this, we exploit the performance of a recent published model Swin UNETR~\cite{tang2022self} on this task. As can be seen, Swin UNETR with self-supervised pre-training on unlabeled data has a 1.7\% improvement against its vanilla version. 
We also investigate the performance of TotalSegmentator~\cite{wasserthal2023totalsegmentator} on this dataset, and find that it falls behind models that are specifically trained on the target dataset.
Our method compares favorably with the average DSC of 89.7\%, which is a competitive result ranking in top 3\% among all 1,162 entries in the validation leaderboard without heavy model ensembles. The workflow of spine parsing and multi-organ segmentation is illustrated in~\cref{fig:vert_organ_vis}.

\subsection{Ablation Study}
We perform ablation study on RibInst to investigate the effectiveness of three key factors in our framework.

\noindent {\bf Effects of the Weighted Adjacency Matrix.} To validate the efficacy of the preset weighted adjacency matrix, we conduct an ablation study w.r.t. $\lambda_m$, the scaling factor of the matrix, which is also known as index cost. Candidate parameter is chosen from $[0, 1, 2, 4, 8]$ where 0 means the vanilla bipartite matching. We should notice that when $\lambda_m$ is large enough, it can be regarded as directly assigning the query to a fixed ground-truth label. \cref{fig:ablation}(a) reveals that a proper value of $\lambda_m = 4$ stands out with the highest Id.Rate of rib labeling. One explanation is that a reasonable perturbation term during the matching process could gradually impose anatomical semantic labels on queries and help build the correct relations among them.

\noindent {\bf Steerable at Inference?} To further validate the steerable attribute of the proposed model, we randomly pick some rib labels with varied number from 4 to 24 with a step size of 4 at inference. The task is to retrieve the pre-selected ribs with query embeddings in the same sequence where Id.Rate for the given subset of ribs is the evaluation metric. 
The purpose of this experiment is to evaluate whether the mapping between the queries and the ground-truth labels has been correctly established. If this mapping is perfectly established, we can use the 1st query embedding to detect the 1st rib, the 2nd query embedding to detect the 2nd rib, and so on.
\cref{fig:ablation}(b) demonstrates the performances of two different settings of with index cost or without index cost. Note that in the round of only 4 ribs are expected to be detected, the model without index cost misses all targets. This is not out of expectation, because the one-to-one relations between query series and ground-truth sequence are not explicitly constructed. 
Take a specific case for instance, suppose we want the model to detect four ribs, with their labels being [1, 2, 3, 4]. We need to examine the detection results of the query embeddings indexed [1, 2, 3, 4] after passing through the model's decoder. In the absence of index cost, if the model outputs four 9-DoF boxes with labels [5, 6, 7, 8], i.e., the 5th to the 8th ribs, the Id.Rate would be 0\%.
As the number of queries increases, the vanilla model can gradually return some correct targets back (although not necessarily detected by the
queries with the same index). On the contrary, Med-Query demonstrates consistent performance when the number of queries varies, despite the fluctuations in the metrics due to changes in the number of targets. This stability enables the realization of our steerable system. In terms of inference efficiency, Med-Query can be further boosted with less queries required, e.g., if 4 queries are required, it can save 17\% of time in detection stage and the ratio can even expand to 50\% in the perspective of full pipeline, as shown in \cref{fig:ablation}(c). The efficiency improvement benefits from less computation in Transformer decoder and segmentation head when fewer queries are executed. During the above ablation experiments, cropped data is not used.

\noindent {\bf Data Augmentation \& Beyond.} \label{sec:ablation_data_aug}We conduct comprehensive experiments to analyze the critical factors in our proposed framework as shown in \cref{tbl:ablation_aug}. The joint operations of \textit{RandomTranslate}, \textit{RandomScale}, and \textit{RandomRotate} in 3D space get the most significant performance gain. \textit{RandomCrop} along the $Z$-axis for imitating limited FOVs in clinical practice can further improve the Id.Rate by 1.8\%. \textit{RandomErase} gets additional performance benefit. We also explore a common trick of imposing relative distance constraints between neighboring center points in landmark detection tasks~\cite{liu2020landmarks}, and obtain a gain of 0.5\%. Additionally, we exploit the efficacy of a 3D extended version of the fixed {\it Sine} spatial positional encoding that has proven to be useful in \cite{carion2020end}. It can be seen that cancellation of this leads to 0.8\% Id.Rate drop if the measurement thresholding is set to (20mm, 20mm, 10$^\circ$). However, if the thresholding is set more strictly, the drop will disappear and even get the opposite results. A similar observation that removing positional encoding only degrades accuracy by a small margin can be found in~\cite{chen2021empirical}. It implies that positional encoding may deserve further investigation especially in 3D scenarios.

\section{Discussion}
\subsection{Technology Limitations}
Although the proposed method has achieved appealing performance on a range of anatomy parsing tasks, there are still several limitations that should be acknowledged. Firstly, our framework requires each stage to be trained separately, rather than being easily end-to-end trainable. This can lead to increased complexity as well as potential difficulties in optimizing the performance of each individual stage. Secondly, despite our careful consideration of the trade-off between efficiency and accuracy, it is difficult to avoid the situation that the final segmentation results are heavily reliant on the accuracy of the detection stage. 
To be more specific, although a 9-DoF box offers a more compact representation compared to an axis-aligned box, it is inevitable that a 9-DoF box for one rib will include a small portion of an adjacent rib due to the close proximity of adjacent ribs. This is generally not problematic, as our segmentation model in the third stage is trained to segment the primary rib within such an ROI. However, as with all detection-then-segmentation approaches, if the box localization is
not very accurate and includes a significant portion of an adjacent rib, it can lead to confusion for the subsequent segmentation model.
Thirdly, for complex abnormal structures present in the ribs, such as the twisted ``X'' shape formed by adjacent ribs, our method may produce inaccurate detection results.
Finally, with regards to steerable inference, our current framework relies on ordered trained query embeddings, which may limit the flexibility and adaptability of the system. In our future work, we will explore the integration of natural language embeddings pre-trained using CLIP~\cite{radford2021learning}, to provide a more intuitive and user-friendly interaction within the network architecture, as in some most recent interactive segmentation methods~\cite{kirillov2023segment, zou2023segment}.

\subsection{Dataset Limitations}
Despite the high quality of our curated RibInst dataset, there are limitations to its generality. It is curated from a dataset that was originally designed for rib fracture detection and classification, meaning that the ribs included in the dataset may have varying degrees of underlying pathologies, limiting their representativeness of the broader (healthy) population. 
Moreover, individuals with atypical numbers of vertebrae and ribs present another challenge for our parsing algorithm. The current model is trained predominantly on standard anatomical datasets, which limits its performance when faced with anatomical variations. Although we employ strategies such as \textit{RandomCrop} and \textit{RandomErase} to simulate limited FOV and fewer anatomical structures than standard situation, our method may struggle to handle cases with 13 pairs of ribs or 6 lumbar vertebrae. This issue could be addressed in the future by collecting more data of such cases or through specialized post-processing techniques.
Therefore, caution must be taken when applying our model to broader real-world scenarios, as it may require additional training and testing on more diverse data distributions. This could include a greater variety of normal/healthy patient CT scans and full-body scans to further ensure its performance and generality.

\section{Conclusion}
In this work, we have presented a steerable, robust and efficient Transformer-based framework for anatomy parsing in CT scans. The pipeline is conducted via following the detection-then-segmentation paradigm and processing input 3D scans at different resolutions progressively. To our best knowledge, this work is the first to estimate the 9-DoF representation for object detection in 3D medical imaging via one-stage Transformer. The resulted method can be executed in a steerable way to directly retrieve any anatomy of interest and further boost the inference efficiency. It is a unified computing framework that can generalize well to a variety of anatomy parsing tasks and performs better or on par with state-of-the-art methods in anatomy instance detection, identification and segmentation. We have released our annotations, code and models, hoping to benefit the community and facilitate the future development on automatic parsing of anatomical structures. 

\bibliographystyle{IEEEtran}
\bibliography{mybib}

\end{document}